\documentclass[10pt,twocolumn,letterpaper]{article}
\usepackage[pagenumbers]{cvpr}

\usepackage{amssymb}
\usepackage{latexsym}
\usepackage{graphicx}

\usepackage{url}
\usepackage{xcolor}

\usepackage{hyperref}
\usepackage{booktabs}

\usepackage{amsmath}
\usepackage{algorithm, algorithmic}

\newcommand{\tworow}[2]{\begin{tabular}{@{}l@{}}#1 \\ #2\end{tabular}}

\DeclareMathOperator*{\amax}{\mathrm{argmax}}

\definecolor{newcolor}{rgb}{.8,.349,.1}

\begin{document}

%\verso{Maciej Mazurowski \textit{et~al.}}

%\begin{frontmatter}

\title{Segment Anything Model for Medical Image Analysis: an Experimental Study}

\author{%
  Maciej A. Mazurowski$^{1,2,3,4}$ \quad Haoyu Dong$^{2}$ \quad Hanxue Gu$^{2}$\\
  Jichen Yang$^{2}$ \quad Nicholas Konz$^{2}$ \quad Yixin Zhang$^{2}$\\
  $^{1}$ Department of Radiology \quad $^{2}$ Department of Electrical and Computer Engineering \quad \\
  $^{3}$ Department of Computer Science \quad $^{4}$ Department of Biostatistics \& Bioinformatics \\
  Duke University, NC, USA\\
  {\tt\small \{maciej.mazurowski, haoyu.dong151, hanxue.gu,}\\
  {\tt\small jichen.yang, nicholas.konz, yixin.zhang7\}@duke.edu} \\
  \textit{Code:} \small{\url{https://github.com/mazurowski-lab/segment-anything-medical-evaluation}}
}
\maketitle

\begin{abstract}
Training segmentation models for medical images continues to be challenging due to the limited availability of data annotations. Segment Anything Model (SAM) is a foundation model that is intended to segment user-defined objects of interest in an interactive manner. While the performance on natural images is impressive, medical image domains pose their own set of challenges. Here, we perform an extensive evaluation of SAM’s ability to segment medical images on a collection of $19$ medical imaging datasets from various modalities and anatomies. We report the following findings: (1) SAM’s performance based on single prompts highly varies depending on the dataset and the task, from IoU=$0.1135$ for spine MRI to IoU=$0.8650$ for hip X-ray. (2) Segmentation performance appears to be better for well-circumscribed objects with prompts with less ambiguity and poorer in various other scenarios such as the segmentation of brain tumors. (3) SAM performs notably better with box prompts than with point prompts. (4) SAM outperforms similar methods RITM, SimpleClick, and FocalClick in almost all single-point prompt settings. (5) When multiple-point prompts are provided iteratively, SAM’s performance generally improves only slightly while other methods' performance improves to the level that surpasses SAM's point-based performance. We also provide several illustrations for SAM's performance on all tested datasets, iterative segmentation, and SAM’s behavior given prompt ambiguity. We conclude that SAM shows impressive zero-shot segmentation performance for certain medical imaging datasets, but moderate to poor performance for others. SAM has the potential to make a significant impact in automated medical image segmentation in medical imaging, but appropriate care needs to be applied when using it.

%%%%
\end{abstract}

%\end{frontmatter}

%\linenumbers

%% main text
\section{Introduction}

Image segmentation is a central task in medical image analysis, ranging from the segmentation of organs \cite{rister_ct-org_2019}, abnormalities \cite{anwar2018medical}, bones \cite{kayalibay2017cnn}, and others \cite{pham2000current}, which has received significant advancements from deep learning \cite{hesamian2019deep, razzak2018deep}. However, developing and training segmentation models for new medical imaging data and/or tasks is practically challenging, due to the expensive and time-consuming nature of collecting and curating medical images, primarily because trained radiologists must typically provide careful mask annotations for images.

These difficulties could be significantly mitigated with the advent of foundation models \cite{zhou2023comprehensive} and zero-shot learning \cite{Chen2021KnowledgeawareZL}. Foundation models are neural networks trained on an extensive amount of data, using creative learning and prompting objectives that typically do not require traditional supervised training labels, both of which contribute towards the ability to perform zero-shot learning on completely new data in a variety of settings. Foundation models have shown paradigm-shifting abilities in the domain of natural language processing \cite{openai2023gpt4}. The recently developed Segment Anything Model is a foundation model that has achieved promising zero-shot segmentation performance on a variety of natural image datasets \cite{SAM}.

\subsection{What is SAM?}
Segment Anything Model (SAM) is designed to segment an object of interest in an image given certain prompts provided by a user. Prompts can take the form of a single point, a set of points (including an entire mask), a bounding box, or text. The model is asked to return a valid segmentation mask even in the presence of ambiguity in the prompt. The general idea behind this approach is that the model has learned the concept of an object and thus can segment any object that is pointed out. This results in a high potential for it to be able to segment objects of types that it has not seen without any additional training, \textit{i.e.}, high performance in the zero-shot learning regime. In addition to the prompt-based definition of the task, the SAM authors utilized a specific model architecture and a uniquely large dataset to achieve this goal, described as follows. 

SAM was trained progressively alongside the development of the dataset of images with corresponding object masks (SA-1B). The dataset was developed in three stages. First, a set of images was annotated by human annotators by clicking on objects and manually refining masks generated by SAM, which at that point was trained on public datasets. Second, the annotators were asked to segment masks that were not confidently generated by SAM to increase the diversity of objects. The final set of masks was generated automatically by prompting the SAM model with a set of points distributed in a grid across the image and selecting confident and stable masks. 

\subsection{How to segment medical images with SAM?}
SAM is designed to require a prompt or a set of prompts to produce a segmentation mask. Technically, the model can be run without a prompt to provide any visible object, but we do not expect this to be useful for medical images, where there are often many other objects in the image beside the one of interest. Given this prompt-based nature, in its basic form, SAM cannot be used the same way as most segmentation models in medical imaging where the input is simply an image and the output is a segmentation mask or multiple masks for the desired object or objects. 

\begin{table*}[ht!]
  \centering
  \begin{tabular}{lllclc}
    \toprule
    \bf \tworow{Abbreviated}{dataset name} & \bf Full dataset name and citation & \bf Modality & \bf \tworow{Num.}{classes} & \bf Object(s) of interest & \bf \tworow{Num.}{masks} \\\midrule
    MRI-Spine & \tworow{Spinal Cord Grey Matter}{Segmentation Challenge \cite{prados2017spinal}} & MRI & 2 & \tworow{Gray matter,}{spinal cord} & 551 \\\midrule
    MRI-Heart & Medical Segmentation Decathlon \cite{simpson2019largeheartMRI} & MRI & 1 & Heart & 1,301 \\\midrule
    MRI-Prostate & \tworow{Initiative for Collaborative}{Computer Vision Benchmarking \cite{lemaitre2015prostateMRI}} & MRI & 1 & Prostate & 893 \\\midrule
    MRI-Brain & \tworow{The Multimodal Brain Tumor Image}{Segmentation Benchmark (BraTS) \cite{menze2014multimodalbrats}} & MRI & 3 & \tworow{GD-enhancing tumor,}{\tworow{Peritumoral edema,}{\tworow{necrotic and non-}{enhancing tumor core}}} & 12,591 \\\midrule
    MRI-Breast & \tworow{Duke Breast Cancer MRI:}{Breast + FGT Segmentation \cite{saha2018machine, hu2022fully}} & MRI & 2 & \tworow{Breast, fibrog-}{landular tissue} & 503 \\\midrule
    
    Xray-Chest & \tworow{Montgomery County and Shenzhen}{Chest X-ray Datasets \cite{jaeger2014chestxray}} & X-ray & 1 & Chest & 704 \\\midrule
    Xray-Hip & X-ray Images of the Hip Joints \cite{gut_x-ray_2021} & X-ray & 2 & Ilium, femur & 140 \\\midrule
    
    US-Breast & Dataset of Breast Ultrasound Images\cite{al2020datasetultrasoundbreast} & Ultrasound & 1 & Breast & 647 \\\midrule
    US-Kidney & CT2US for Kidney Segmentation \cite{song2022ct2us} & Ultrasound & 1 & Kidney & 4,586 \\\midrule
    US-Muscle & \tworow{Transverse Musculoskeletal}{Ultrasound Image Segmentations \cite{marzola2021deep}} & Ultrasound & 1 & Muscle & 4,044 \\\midrule
    US-Nerve & \tworow{Ultrasound Nerve Segmentation }
    {Identify (\cite{ultrasound-nerve-segmentation})} & Ultrasound & 1 & Nerve & 2,323 \\\midrule
    US-Ovarian-Tumor & \tworow{Multi-Modality Ovarian Tumor}{Ultrasound (MMOTU) \cite{zhao2022multi}} & Ultrasound & 1 & Ovarian tumor & 1,469 \\\midrule

\end{tabular}
\caption{\textbf{All datasets evaluated in this paper. ``num. masks'' refers to the number of images with non-zero masks. For 3D modalities, 2D slices are used as inputs.}}
\label{tab:datasets1}
\end{table*}

\begin{table*}[ht]
  \centering
  \begin{tabular}{lllclc}
    \toprule
    \bf \tworow{Abbreviated}{dataset name} & \bf Full dataset name and citation & \bf Modality & \bf \tworow{Num.}{classes} & \bf Object(s) of interest & \bf \tworow{Num.}{masks} \\\midrule
    CT-Colon & Medical Segmentation Decathlon \cite{simpson2019largeheartMRI} & CT & 1 & \tworow{Colon cancer}{primaries} & 1,285 \\\midrule
    CT-HepaticVessel & Medical Segmentation Decathlon \cite{simpson2019largeheartMRI} & CT & 1 & Vessels, tumors & 13,046 \\\midrule
    CT-Pancreas & Medical Segmentation Decathlon \cite{simpson2019largeheartMRI} & CT & 1 & \tworow{parenchyma}{and mass} & 8,792 \\\midrule
    CT-Spleen & Medical Segmentation Decathlon \cite{simpson2019largeheartMRI} & CT & 1 & spleen & 1,051 \\\midrule
    CT-Liver & \tworow{The Liver Tumor}{Segmentation Benchmark (LiTS) \cite{bilic2023lits}} & CT & 1 & Liver & 5,501 \\\midrule
    CT-Organ & \tworow{CT Volumes with Multiple}{Organ Segmentations (CT-ORG) \cite{rister_ct-org_2019}} & CT & 5 & \tworow{Liver, bladder, lungs,}{kidney, bone} & 4,776 \\\midrule    
    PET-Whole-Body & \tworow{A FDG-PET/CT dataset}{with annotated tumor lesions \cite{gatidis2022whole}} & PET/CT & 1 & Lesion & 1,015 \\    \midrule 
  \end{tabular}
  \caption{\textbf{(Continued) All datasets evaluated in this paper. ``num. masks'' refers to the number of images with non-zero masks. For 3D modalities, 2D slices are used as inputs.}}
  \label{tab:datasets2}
\end{table*}

We propose that there are three main ways in which SAM can be used in the process of segmentation of medical images. 
The first two involve using the actual Segment Anything Model in the process of annotation, mask generation, or training of additional models. These approaches do not involve changes to the SAM . The third approach involves the process of training/fine-tuning a SAM-like model targeted for medical images. We detail each approach next. Note that we do not comment here on text-based prompting, as it is still in the proof-of-concept stage for SAM. 

\noindent\textbf{Semi-automated annotation (“human in the loop”).} 
The manual annotation of medical images is one of the main challenges of developing segmentation models in this field since it typically requires the valuable time of physicians. SAM could be used in this setting as a tool for faster annotation. This could be done in different ways. In the simplest case, a human user provides prompts for SAM, which generates a mask to be approved or modified by the user; this could be refined iteratively. Another option is where SAM is given prompts distributed in a grid across the image (the “segment everything” mode), and generates masks for multiple objects which are then named, selected, and/or modified by the user. This is only the start; many other possibilities could be imagined.

\noindent\textbf{SAM assisting other segmentation models.} 
One version of this usage mode is where SAM works alongside another algorithm to automatically segment images (an “inference mode”). For example, SAM, based on point prompts distributed across the image, could generate multiple object masks which then could be classified as specific objects by a separate classification model. Similarly, an independent detection model, \textit{e.g.,} ViTDet \cite{Li2022ExploringPV}, could generate object-bounding boxes of images to be used as prompts for SAM to generate precise segmentation masks. 

Furthermore, SAM could be used in the training loop of some other semantic segmentation model. For example, the masks generated by a segmentation model on unlabeled images during training could be used as prompts to SAM to generate more precise masks for these images, which could be used as iteratively refined supervised training examples for the model being trained. One could conceptualize many other specific modes of including SAM in the process of training new segmentation models.

\noindent\textbf{New medical image foundation segmentation models.} 
In this usage mode, the development process of a new segmentation foundation model for medical images could be guided by SAM’s own development process. The largest difficulty of this would be in the much lower availability of medical images and quality annotations, compared to natural images, but this is possible in principle. A more feasible option could be to fine-tune SAM on medical images and masks from a variety of medical imaging domains, rather than training from scratch, as this would likely require fewer images.

\section{Methodology}
In the previous section, we described various usage scenarios of SAM for medical image segmentation. These are conceptually promising but largely rely on the assumption that SAM can generate accurate segmentations of medical images. Here, we experimentally evaluate this claim and evaluate the performance of SAM within a variety of different realistic usage scenarios and datasets in medical imaging.

\begin{algorithm}[h!]
    \caption{Prompt Point Generation Scheme}
    \label{alg:promptgen}
    \textbf{Input}: Image $X\in\mathbb{R}^{H\times W}$, ground truth mask $M\in\{0,1\}^{H\times W}$, Segment Anything Model \texttt{SAM}, prompt count $N$, closest-zero-pixel distance function $d=\texttt{distanceTransform()}$ from OpenCV \cite{opencv_library}.\\
    \begin{algorithmic}[1] %[1] enables line numbers
    \STATE \textit{Initialize first prompt point $p_1$ as the point within the mask foreground farthest from the background:}% $p_{ij}$ within $M_{ij}=1$ {\tt == 1} farthest from $M$'s boundary:
    \STATE $\mathcal{P}={\amax\limits_{(i,j)}}(d[(i, j), (k,l)])$ for all $(i,j), (k,l)$ such that $M_{ij}=1$, $M_{kl}=0$.
    \STATE \textit{Choose randomly if multiple points satisfy this:}
    \STATE $p_1=\texttt{random\_choice}(\mathcal{P})$
    \STATE \textit{Predict mask} $Y_1= \texttt{SAM}(X, p_1)$

    \STATE \textit{Get prediction error region} $E_1 = Y_1\cup M - Y_1\cap M$
    \STATE \textit{Subsequent prompt points are those farthest from the boundary of iteratively-updating error region $E_n$:}
    \FOR{$n = 2, \ldots , N$}
        \STATE $\mathcal{P}_n={\amax\limits_{(i,j)}}(d[(i, j), (k,l)])$ for all $(i,j), (k,l)$ such that $[E_{n-1}]_{ij}=1$, $[E_{n-1}]_{kl}=0$.
        \STATE $p_n=\texttt{random\_choice}(\mathcal{P}_n)$
        \STATE $Y_n= \texttt{SAM}(X, p_n)$
        \STATE $E_n = Y_n\cup M - Y_n\cap M$
    \ENDFOR
    \RETURN Prompt points $p_1, \ldots, p_N\in \mathbb{N}^2$
    \end{algorithmic}
\end{algorithm}

\subsection{Datasets}
We compiled and curated a set of 19 publicly available medical imaging datasets for image segmentation. While the phrase “medical imaging” is sometimes used to refer to all images pertaining to medicine, we focus on the common definition of  radiological images. Our dataset includes planar X-rays, magnetic resonance images (MRIs), computed tomography (CT) images, ultrasound (US) images, and positron emission tomography (PET) images. The datasets are summarized in Tables \ref{tab:datasets1} and \ref{tab:datasets2}. For datasets containing more than one type of object, we considered the segmentation of each object as a separate task.

\subsection{Experiments}
We performed a thorough evaluation of SAM with both non-iterative prompts (generated prior to SAM being applied) and iterative prompts (generated after seeing the model's predictions). We also explored the “segment everything” mode of SAM and analyzed the different outputs that SAM generates in response to ambiguity in the prompts.

\begin{figure*}[ht!]
    \centering
    \includegraphics[width=1\linewidth]{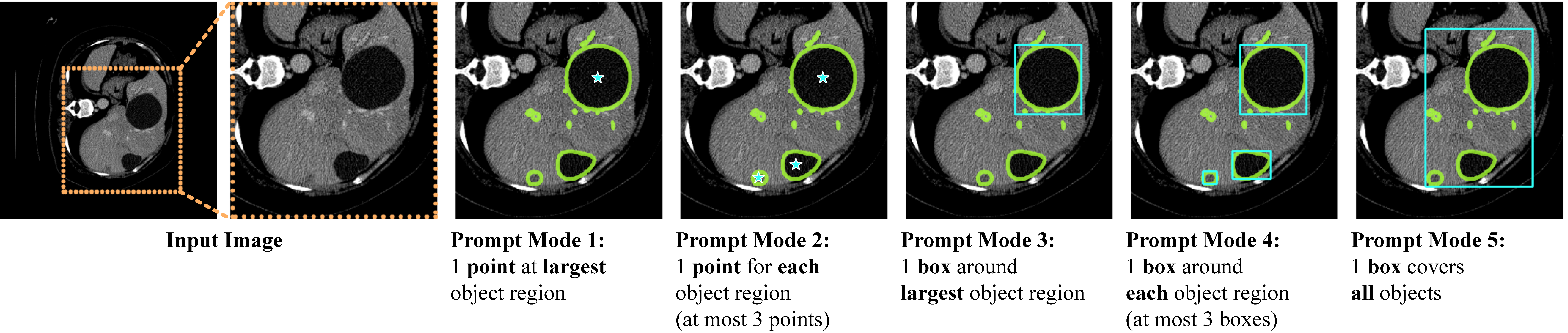}
    \caption{Examples of prompt(s) generated by the five modes respectively. Green contours show the ground-truth masks, and blue star(s) and box(es) indicate the prompts.}
    \label{fig:prompt mode}
\end{figure*}
\begin{figure*}[t]
    \centering
    \includegraphics[width=1\linewidth]{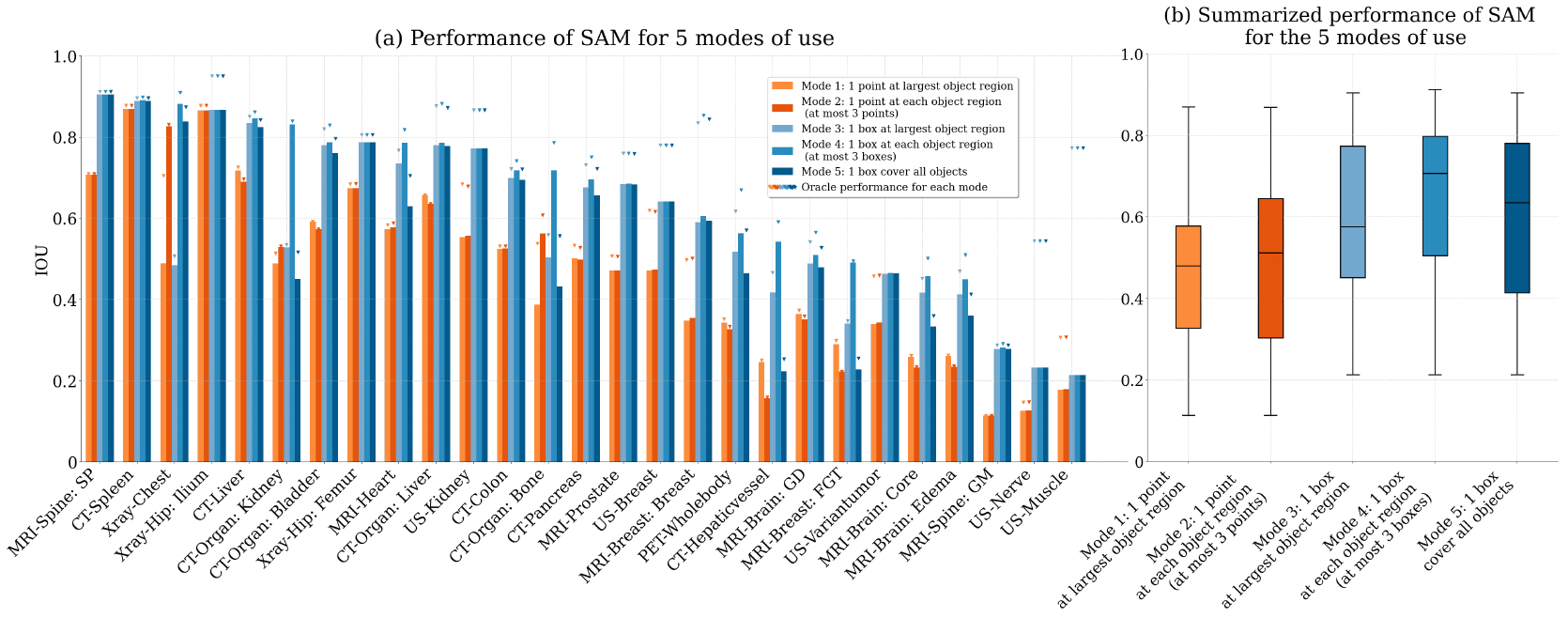}
    \caption{Performance of SAM under 5 modes of use. Left: Performance of SAM across 28 segmentation tasks, with results ranked in descending order based on Mode 4. Oracle performance for each mode is indicated by the inverted triangle. Right: A summarized performance comparison of all five modes across all tasks, presented in a box and whisker plot format.}
    \label{fig:five-mode}
\end{figure*}

\begin{figure*}[t]
    \centering
    \includegraphics[width=0.85\linewidth]{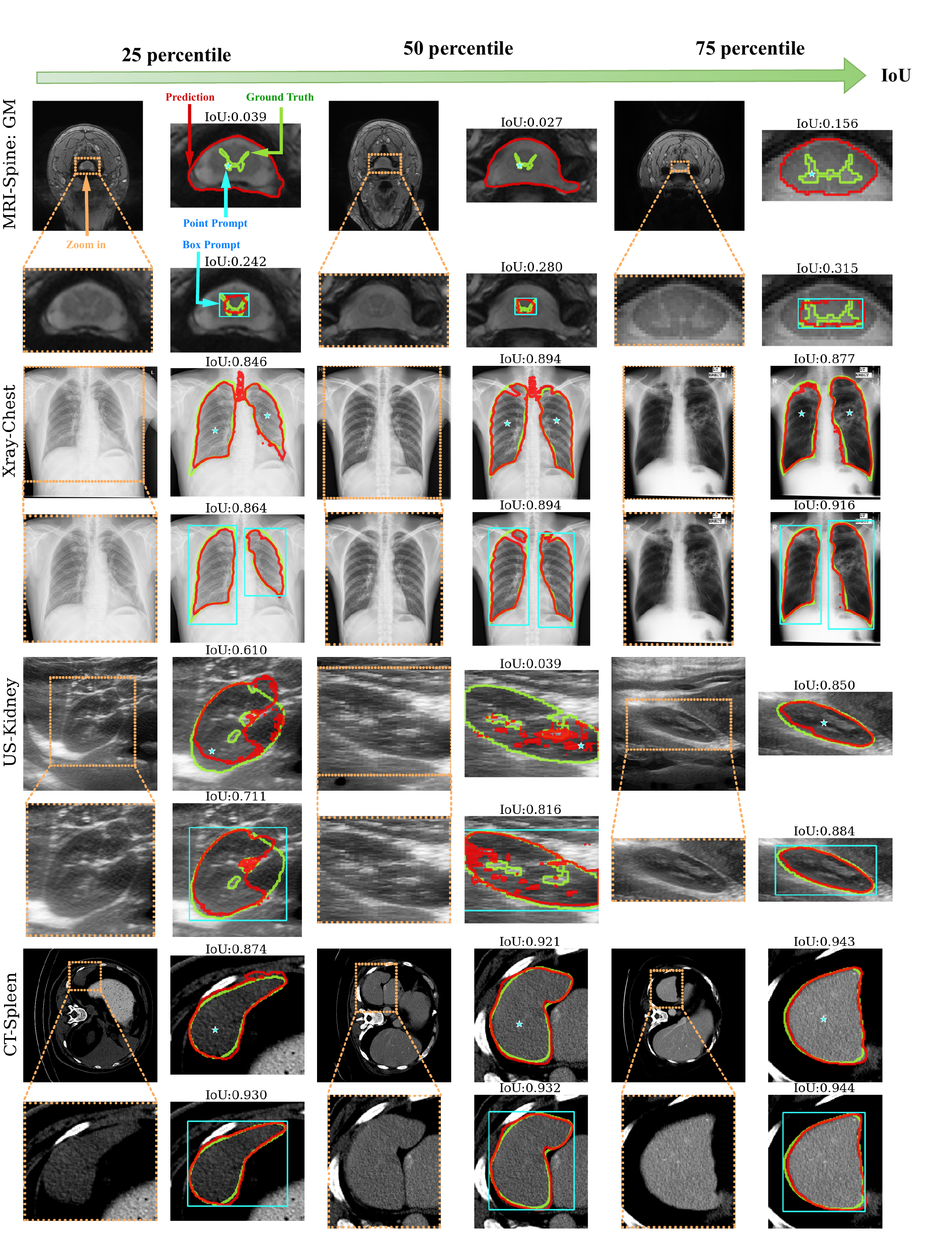}
    \caption{Visualization of SAM's segmentation results in two different modes. Each dataset is shown in two sequential rows, with its name along the left side. For each dataset, it displays three examples from left to right, reflecting the 25th, 50th, and 75th percentiles of IoU across all images for that dataset. For each example, we visualize (top left) the raw image; (bottom left) the zoom-in image with the area of interest; (top right) the segmented results for mode 2: 1 point at each object region; (bottom right) the segmented results for mode 4: 1 box region at each object region. Additionally, the IoU is represented above each segmented result. Examples of all the datasets are shown in Appendix Figure 1-5.}
    \label{fig:vis-mode2-mode4}
\end{figure*}

\subsubsection{Prompting Strategies}
\label{sec:ps}

\noindent\textbf{Non-iteratvie prompts.}
In this primary mode of evaluation, prompts were simulated to reflect how a human user might generate them while looking at the objects. We focus on five modes of non-iterative prompting designed to capture the realistic usage cases of SAM for generating image masks, using either points or bounding boxes. An essential thing to consider is that a single ``object'' of interest / ``ground truth'' mask may consist of multiple disconnected parts, which is especially common in medical images. An example of this is a cross-sectional image of a liver (such as MRI or CT) where in one 3D slice, the liver is portrayed as two non-contiguous areas. Given this consideration, we introduce the following five prompting modes: 
\begin{itemize}
    \item One prompt point is placed at the center of the \textbf{largest} contiguous region of the object of interest/ground truth mask.
    \item A prompt point is placed at the center of \textbf{each} separate contiguous region of the object of interest (up to three points).
    \item One box prompt is placed to tightly enclose the \textbf{largest} contiguous region of the object of interest.
    \item A box prompt is placed to tightly enclose \textbf{each} separate contiguous region of the object of interest (up to three boxes)
    \item A single box is placed to tightly enclose the \textbf{entire} object mask.
\end{itemize}

Juxtaposed examples of each prompting mode for the same image are shown in Figure \ref{fig:prompt mode}. Modes 1 and 2 are equivalent if the object consists of only one contiguous region/part, and the same is true for modes 3, 4, and 5. The point prompts were generated as the point farthest from the boundary of the mask for the object or its part.

\noindent\textbf{Iterative prompts.}
We use a common, intuitive strategy for simulating realistic iterative point prompts, which reflects how those could be generated by a user in an interactive way \cite{Mahadevan2018IterativelyTI}. The details of the prompt generation are illustrated in Algorithm \ref{alg:promptgen}. Specifically, once the network makes a prediction, we compute an error map where both false positive and false negative predictions are marked as 1, \textit{i.e.,} the point furthest from 0. Then we can find the location of the next prompt as the central location of the largest component of the error mask. The label of the prompt is based on whether the new location is foreground or background.

\begin{figure*}[ht!]
    \centering
    \includegraphics[width=1\linewidth]{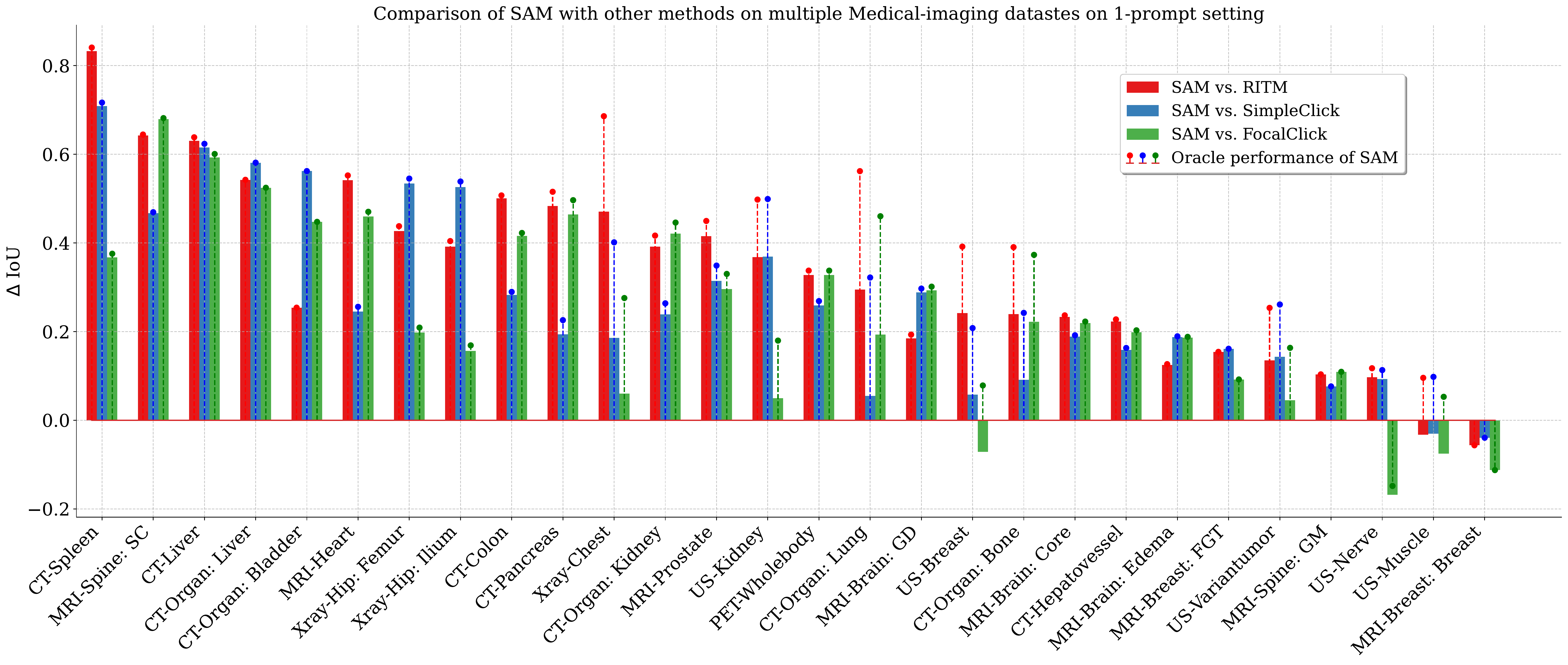}
    \caption{Comparison of SAM with three other competing methods, namely RITM, SimpleClick, and Focalclick, under the 1-point prompt setting. The results are presented in the form of the difference between SAM and other methods ($\Delta$ IoU), and ranked based on the descending order of the largest $\Delta$ IoU for each task.}
    \label{fig:1-point}
\end{figure*}

\begin{figure*}[t!]
    \centering
    \includegraphics[width=1\linewidth]{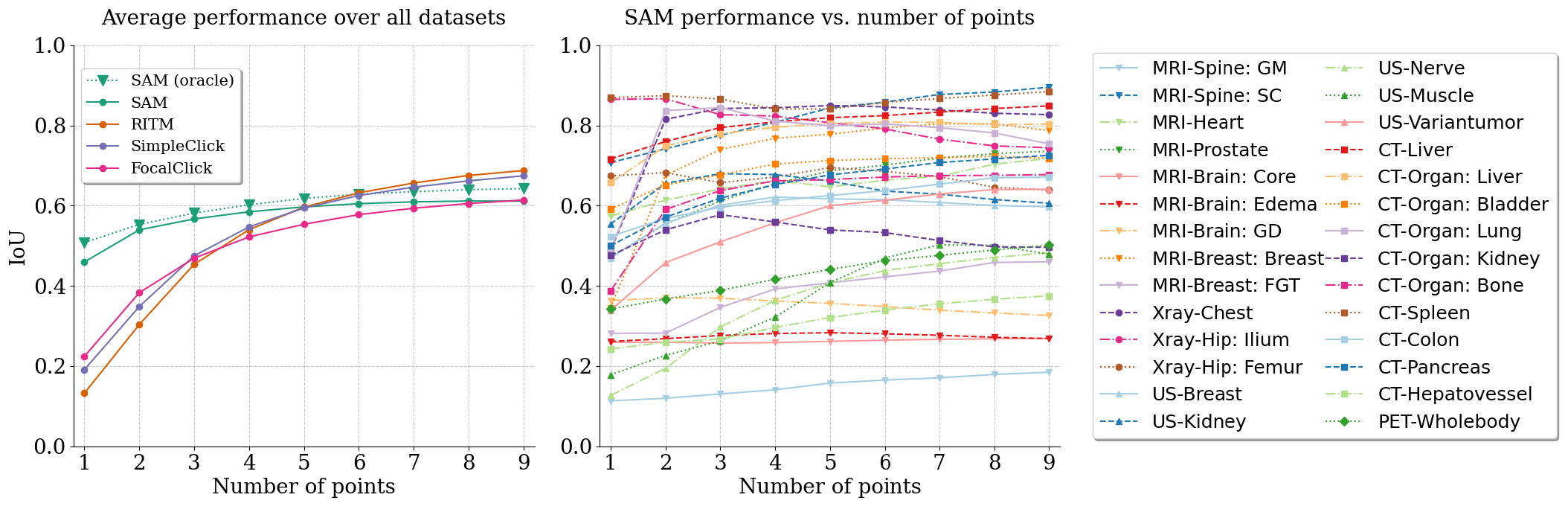}
    \caption{Comparison of SAM and other methods under an interactive prompt setting. (Left) it presents the average performance of SAM and other methods across all tasks with respect to the number of prompt changes. 
    (Right) it shows the detailed performance of SAM over each task.}
    \label{fig:methods}
\end{figure*}

\noindent\textbf{Prompt ambiguity and oracle performance.}
Prompts can be ambiguous in the sense that it may be unclear which object in the image the prompt is referring to. A typical scenario is when objects are nested within each other in the image. 
For example, when a user provides a point prompt within a necrotic component of a brain tumor, they could intend to segment that component, the entire tumor, one hemisphere of the brain, the entire brain, or the entire head. In response to this issue, SAM provides multiple outputs aiming at disambiguating the prompts. This is a very important and practical feature of SAM since in the interactive segmentation setting, multiple potential outputs could be presented to the user, from which they could select the one which is the closest to the object that they intended. In our experiments, we display some examples of the multiple outputs generated by SAM to illustrate how it deals with the ambiguity of the prompts.

Related to the ambiguity, in all experiments, we also present what the developers of SAM call “oracle performance”. This is the performance of the model when the prediction closest (in terms of IoU) to the true mask, \textit{i.e.}, oracle prediction, is always used, out of SAM’s three generated predictions. 
Note that this prediction may differ from SAM’s most confident prediction. While this assumes knowledge of the true mask and is a biased way to assess performance when there is no additional interaction with the user after providing the initial prompts, it is a practical reflection of performance in a setting where the user can select one of the masks generated by SAM. When prompts are generated iteratively, the oracle prediction is used to create the error map that guides the location of the next prompt.

\begin{figure*}[t]
    \centering
    \includegraphics[width=0.97\linewidth]{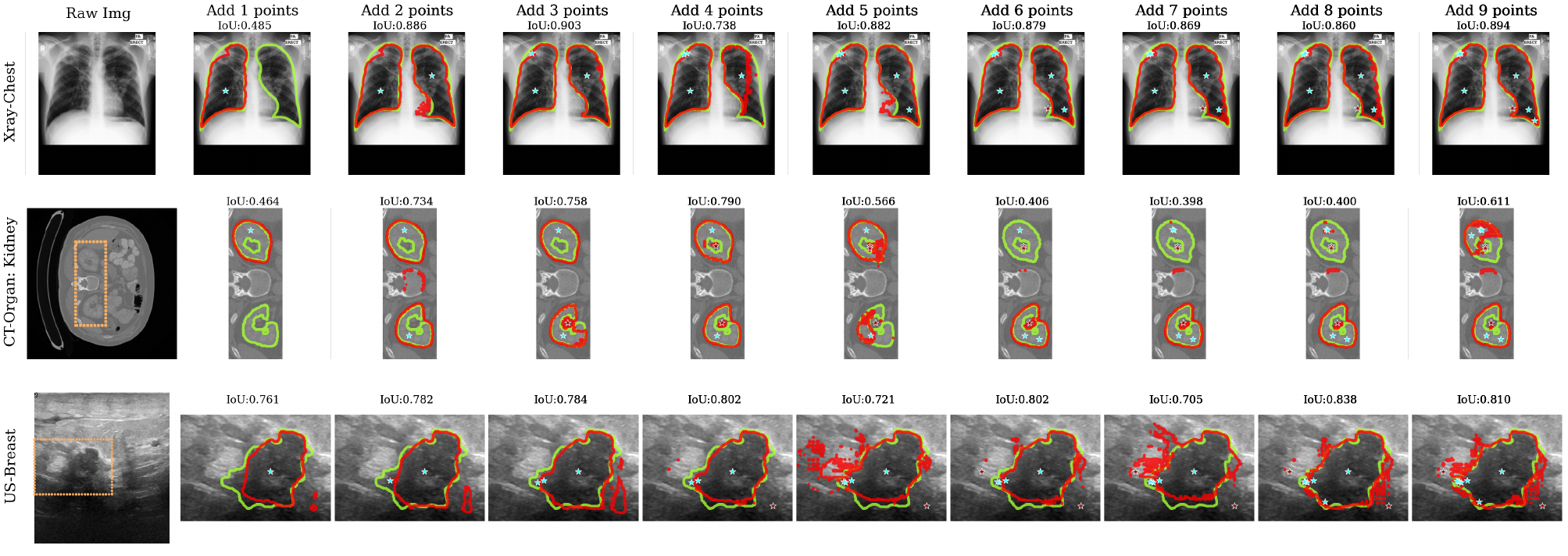}
    \caption{Examples of SAM's prediction under the interactive prompt setting. For each dataset, we display the results from 1-point prompts to 9-point prompts, respectively. The positive prompts are represented as green stars, and the negative prompts are represented as red stars.}
    \label{fig:multi-prmopts}
\end{figure*}

\begin{figure*}[h!]
    \centering
    \includegraphics[width=0.97\linewidth]{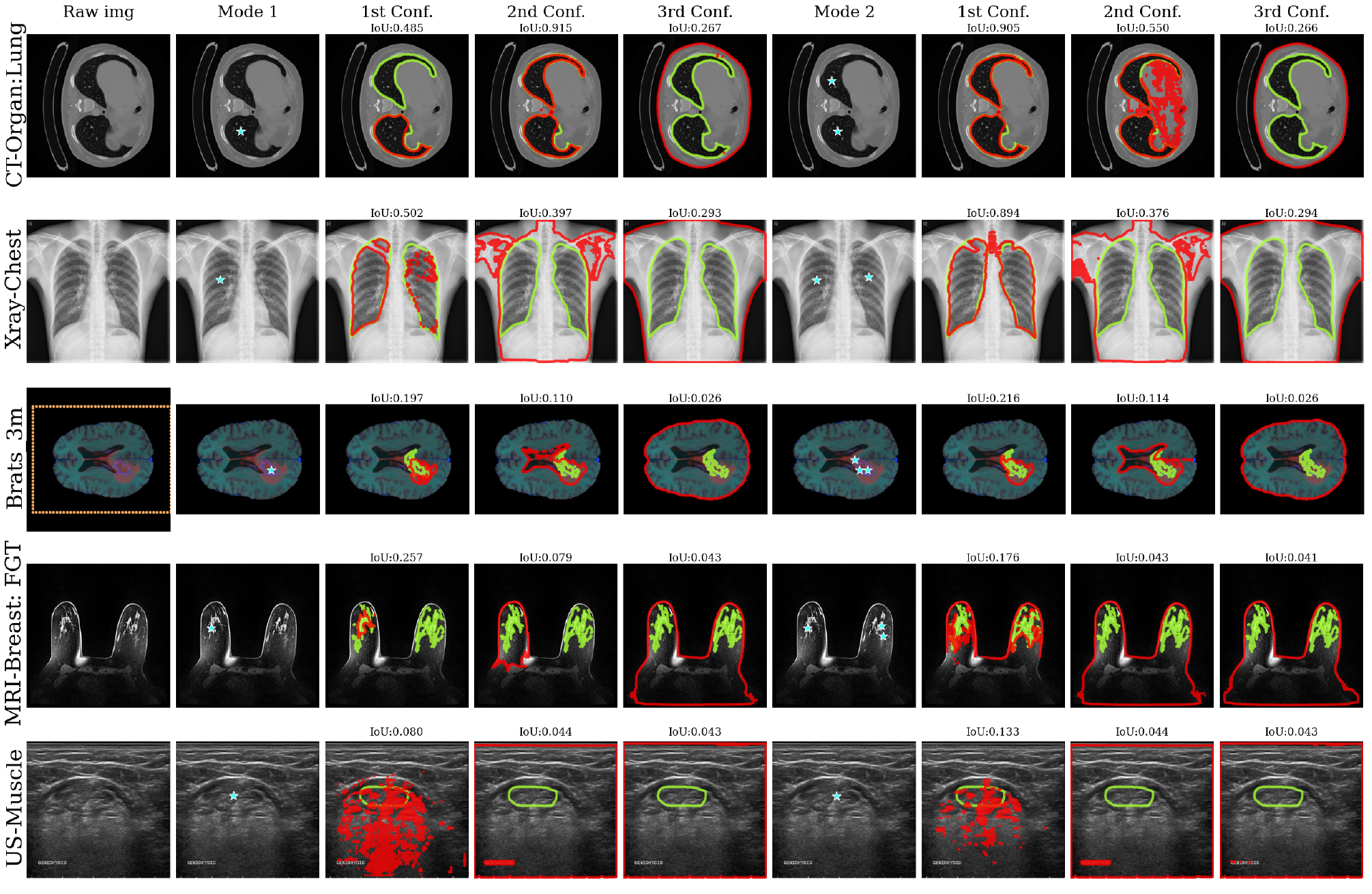}
    \caption{Visualizations of examples with ambiguity based on SAM; the 1st, 2nd, and 3rd confident predictions are shown sequentially.}
    \label{fig:ambi}
\end{figure*}

\subsubsection{Comparison with other methods}
We compared SAM with three interactive segmentation methods, namely RITM \cite{Sofiiuk2021RevivingIT}, SimpleClick \cite{Liu2022SimpleClickII}, and FocalClick \cite{Chen_2022_CVPR}.
RITM adopted HRNet-18 \cite{WangSCJDZLMTWLX19} as the backbone segmentation model and an iterative mask correction approach based on the $\{$ previous mask, new click $\}$ set to achieve outstanding performance on multiple natural imaging datasets. Based on the framework proposed by RITM, SimpleClick replaced the HRNet-18 backbone in RITM with a plain-ViT \cite{dosovitskiy2020vit} and included a ``click embedding'' layer symmetric to patch embedding. It allows for fewer clicks than RITM to achieve an above-threshold performance. FocalClick replaced the HRNet-18 backbone in RITM with a SegFormer \cite{xie2021segformer} and restricted the update responding to new clicks to occur only as local. The method proposed ``progressive merge'' that can exploit morphological information and prevent unintended changes far away from users' clicks, resulting in faster inference time.

\subsubsection{Performance evaluation metric}
For each dataset, we evaluated the accuracy of the masks that SAM and aforementioned methods generate given prompts, with respect to the “ground truth” mask annotations for the given dataset and task. In the quantitative evaluation, we always use the mask with the highest confidence generated by SAM for a given prompt. We used IoU as the evaluation metric, similar to as in SAM’s original paper \cite{SAM}. For datasets containing multiple types of objects, performance was reported independently for each type of object.

\section{Results}
\subsection{Performance of SAM for different modes of use for 28 tasks}
The performance of SAM for our five prompting modes, introduced in Section \ref{sec:ps}, of use is shown in Figure \ref{fig:five-mode}. We draw several conclusions. First, SAMs performance varies widely across different datasets. It ranges from an impressive IoU of $0.9118$ to a very poor IoU of $0.1136$.

\begin{figure*}[t]
    \centering
    \includegraphics[width=0.95\linewidth]{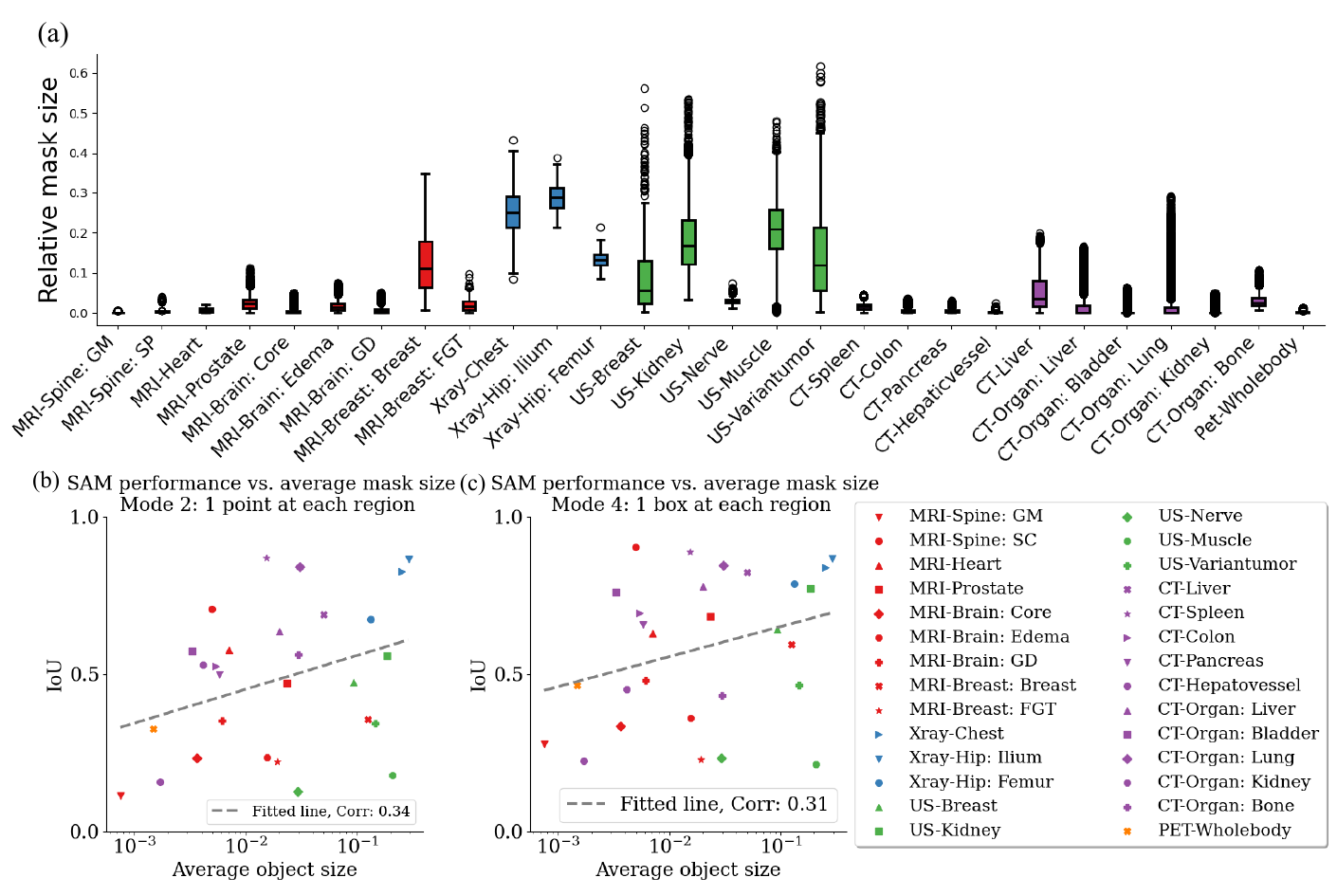}
    \caption{(Top) the relative size of the object in each dataset; (Bottom) the object size vs. detection performance for mode 2 and mode 4 separately; we also show a regression fitted curve each.}
    \label{fig:size}
\end{figure*}

\begin{figure}[ht]
    \centering
    \includegraphics[width=0.95\columnwidth]{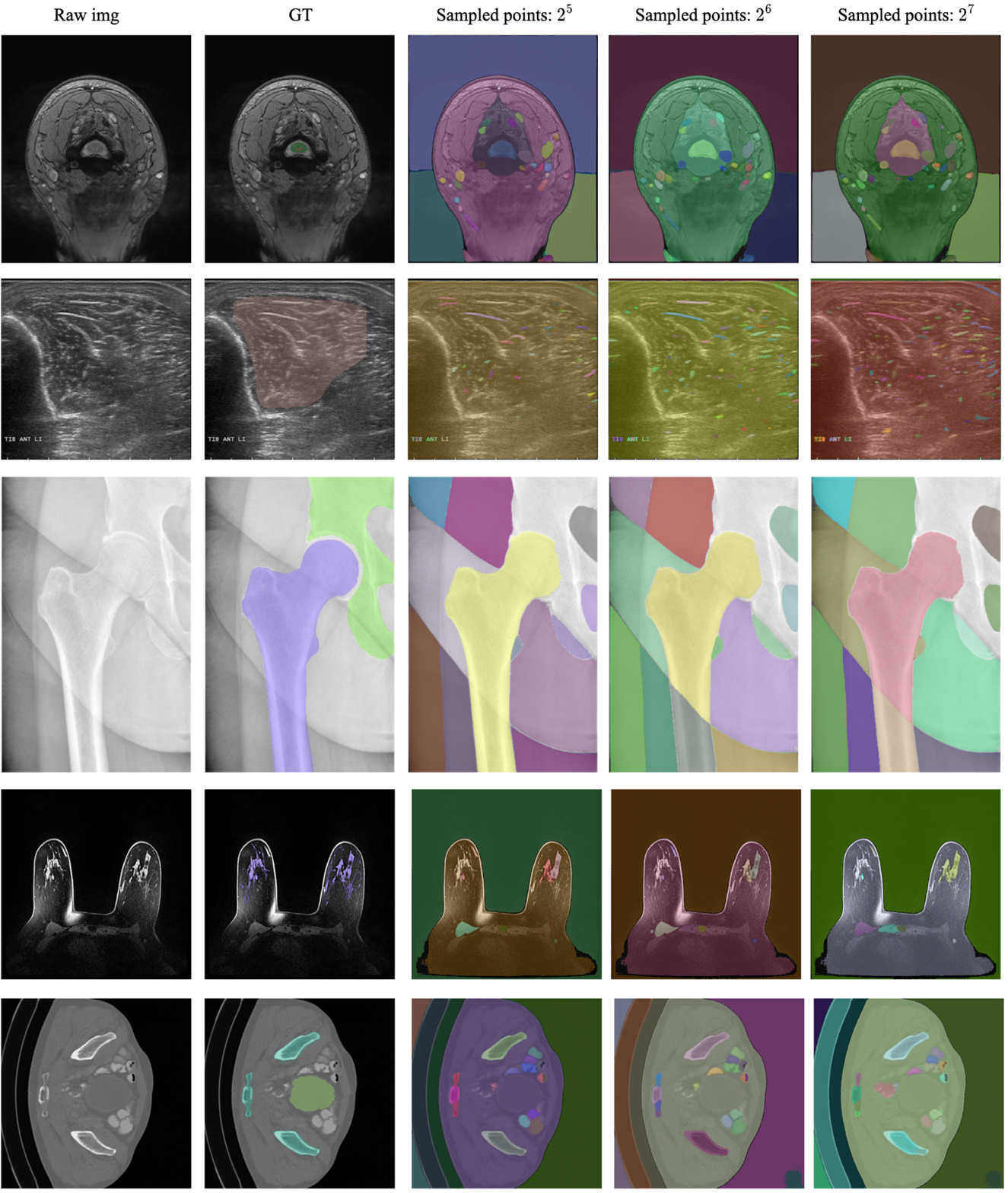}
    \caption{Examples of segment everything mode. For each example, we sampled a different number of grid points at each side as $2^5$,$2^6$ and $2^7$.}
    \label{fig:seg-everything}
\end{figure}

Comparing the performance for different prompting modes shows clear superiority of box prompts over point prompts. Moreover, as expected, prompts where each separate part of the object is indicated separately are generally superior to those where only one part is indicated, or all parts are outlined in one box. This was particularly pronounced for datasets where objects typically consist of more than one part.
Following these two trends, Mode 4, where each part of an object is indicated by a separate box, showed the best performance with an average IoU of $0.6542$.

Additionally, the oracle mode showed a moderate improvement over the default mode. The magnitude of this improvement was highly dependent on the dataset.

Figure \ref{fig:vis-mode2-mode4} shows examples of segmentations generated in prompting Mode 2 (a point for each object part) and Mode 4 (a box around each object part) for 4 selected datasets. For each dataset, we provide examples of SAM’s segmentations in the 25th, 50th, and 75th percentile of IoU. Green contours represent the ground truth masks, red contours represent SAM’s prediction, and teal points or boxes represent the prompts given to SAM. This figure illustrates a high variability of SAM’s performance ranging from near-perfect for well-circumscribed objects with unambiguous prompts to very poor, particularly for objects with ambiguous prompts. A similar illustration for all datasets is provided in Appendix Section B.

\subsection{Comparing SAM to other interactive segmentation methods}
We compare SAM to other interactive methods in the non-iterative prompting setting in Figure \ref{fig:1-point}. SAM performed better than all other methods on $24$ out of $28$ tasks, with a dramatic improvement in performance for some of them. When used in oracle mode, SAM was better than all other methods for $26$ out of $28$ tasks. The average performance across different datasets was $0.4595$ IoU for SAM, $0.5137$ IoU for SAM in oracle mode, $0.2240$ IoU for FocalClick, $0.1910$ IoU for SimpleClick, and $0.1322$ IoU for RITM. Note that SAM exhibits notably better overall performance than all other methods even though it is not used in its best-performing mode of using boxes as prompts, as we used point prompts to have a fair comparison between different methods. If SAM used Mode 3 (single box), it outperformed all other methods in all but one task. The average performance for Mode 3 was $0.5891$ IoU.

\subsection{Performance of SAM and other methods for iterative segmentation}
We show the average performances of SAM, RITM, SimpleClick, and FocalClick in Figure \ref{fig:methods}. 
The detailed performance of the three competing methods over 28 tasks under the interactive prompt setting is shown in Appendix Section C.
As seen in the results above, for the scenario where prompts are provided prior to any interaction with the output of the models, SAM performs notably better than all other methods. However, when additional clicks are provided iteratively with the goal of refining the segmentations returned by the models, the superiority of SAM diminishes, and it is surpassed by two other methods (SimpleClick and RITM) for five or more points provided by the users. This is due to the fact that SAM does not appear to draw hardly any benefits from additional information provided through the interactive points after two or three points have been provided. Figure \ref{fig:multi-prmopts} illustrates this phenomenon. 

We also see that SAM, when given a single prompt, has difficulty segmenting objects with multiple non-contiguous regions. It is more likely to segment one contiguous region instead of trying to find additional semantically similar regions in the entire image. Therefore, in the scenario where multiple regions of interest exist for an object, additional prompt points for SAM can be beneficial if they target additional regions, but beyond that, the benefit of further prompt points is negligible and in some cases such additional input is detrimental.

\subsection{Performance of SAM in the presence of ambiguity of prompts}
When applying SAM on medical images, we found a consistent tendency of SAM in interpreting the point prompts.
As shown in Figure \ref{fig:ambi}, the SAM’s highest-confidence map predictions tend to look similar to results generated by region-growing-based algorithms.
In most cases, a connected region with similar intensity in the image bounded by regions with dissimilar intensities would be segmented. On the other hand, we found masks predictions generated with lower confidence scores may expand  the highest-confidence predictions and tend to have more variety in intensity/texture.

\subsection{Performance of SAM for objects of different sizes}
In Figure \ref{fig:size}(a) we show how object size relates to the performance of SAM. We describe object size as the ratio of the number of pixels in the object to the total in the image. We found object size to vary broadly across our different datasets, by up to over two orders of magnitude. Some datasets also showed high variability of object size within the dataset itself (e.g., kidneys in ultrasound images) while others showed relative consistency of object size (such as bones in hip X-ray images).

Figures \ref{fig:size}(b) and \ref{fig:size}(c) show the relationship between the average object size in a dataset and the performance of SAM. While the correlations are low, there is a trend towards a higher performance of SAM for larger objects. Note that the correlation analysis looks at average object sizes in the image and does not consider the variation of object sizes in the individual datasets. 

\subsection{Segment-everything mode for medical images}
In Figure \ref{fig:seg-everything}, 
we provide an example of using SAM’s “segment everything mode” on medical images. Segment-everything mode uses a dense evenly-spaced mesh of prompt points, designed to direct SAM to segment the image into many different regions. We find that this mode provides mixed usefulness and that the results are somewhat dependent on the number of prompts. While imperfect, this setup could potentially be useful in some applications, \textit{e.g.}, as a ``starting point''  for creating segmentation annotations for different objects in a new image. Selecting masks from the output of segment everything mode is effectively a special case of the prompting mode 2 in our experiments except with additional post-processing. We hence would not elaborate on this mode in our paper.

\section{Conclusions and discussion}
In this study, we evaluated the new Segment Anything Model for the segmentation of medical images. We reached the following conclusions:

\begin{itemize}
\item 	SAM’s accuracy for zero-shot medical image segmentation is moderate on average and varies significantly across different datasets and different images within a dataset. 
\item 	The model performs best with box prompts, particularly when one box is provided for each separate part of the object of interest.
\item 	SAM outperforms RITM, SimpleClick, and FocalClick in the vast majority of the evaluated settings where a single non-iterative prompt point is provided.
\item 	In the setting where multiple iteratively-refined point prompts are provided, SAM obtains very limited benefit from additional point prompts, except for objects with multiple parts. On the other hand, the other algorithms improve notably with additional point prompts, to the level of surpassing SAM’s performance. However, the point prompting modes are inferior to SAM’s box prompting modes.
\item 	We find a small but non-statistically significant correlation between the average object size in a dataset and SAM performance.
\end{itemize}

One of the contributions of our study is that we identified five different modes of use for interactive segmentation methods. This is of particular importance for models which have multiple components, a common feature in medical imaging. These modes also showed different performances in such scenarios. While these modes demonstrate the variety of uses, future work could focus on prompt engineering, both non-iterative and iterative, which could potentially reach even higher performance. 

The segment anything model, associated preprint, and the code illustrate the strengths of open science and the activity of machine learning and machine learning in medical imaging communities. Since we made the first version of this manuscript available, approximately 2 weeks after the release of the SAM paper (or even some before our release), multiple preprints have appeared evaluating SAM in broadly understood medical imaging and radiological imaging \cite{Deng2023SegmentAM, huang2023segment}. Some preprints already showed extensions of SAM to medical imaging \cite{ma2023segment, he2023accuracy} and one paper showed integration of SAM into 3D Slicer \cite{Liu2023SAMMA}. This demonstrates a high likelihood that SAM will become an important part of image segmentation in medical imaging. 

Overall, SAM shows promise for use in medical images, as long as suitable prompting strategies are used for the dataset and task of choice. Future work will include the development of different ways to adapt it to construct medical imaging-specific models as well as extend to 3D segmentation.

\section*{Acknowledgments}
Research reported in this publication was supported by the National Institute Of Biomedical Imaging And Bioengineering of the National Institutes of Health under Award Number R01EB031575 and by the National Heart Lung and Blood Institute of the National Institutes of Health under Award Number R44HL152825. The content is solely the responsibility of the authors and does not necessarily represent the official views of the National Institutes of Health.

%%Harvard
{\small
\bibliographystyle{ieee_fullname}
\bibliography{refs}
}

% Supplementary material that may be helpful in the review process should
% be prepared and provided as a separate electronic file. That file can
% then be transformed into PDF format and submitted along with the
% manuscript and graphic files to the appropriate editorial office.
\section*{Supplementary}
\subsection*{A. Graphical Abstract}
\begin{figure*}
    \centering
    \includegraphics[angle=90, width=1\linewidth]{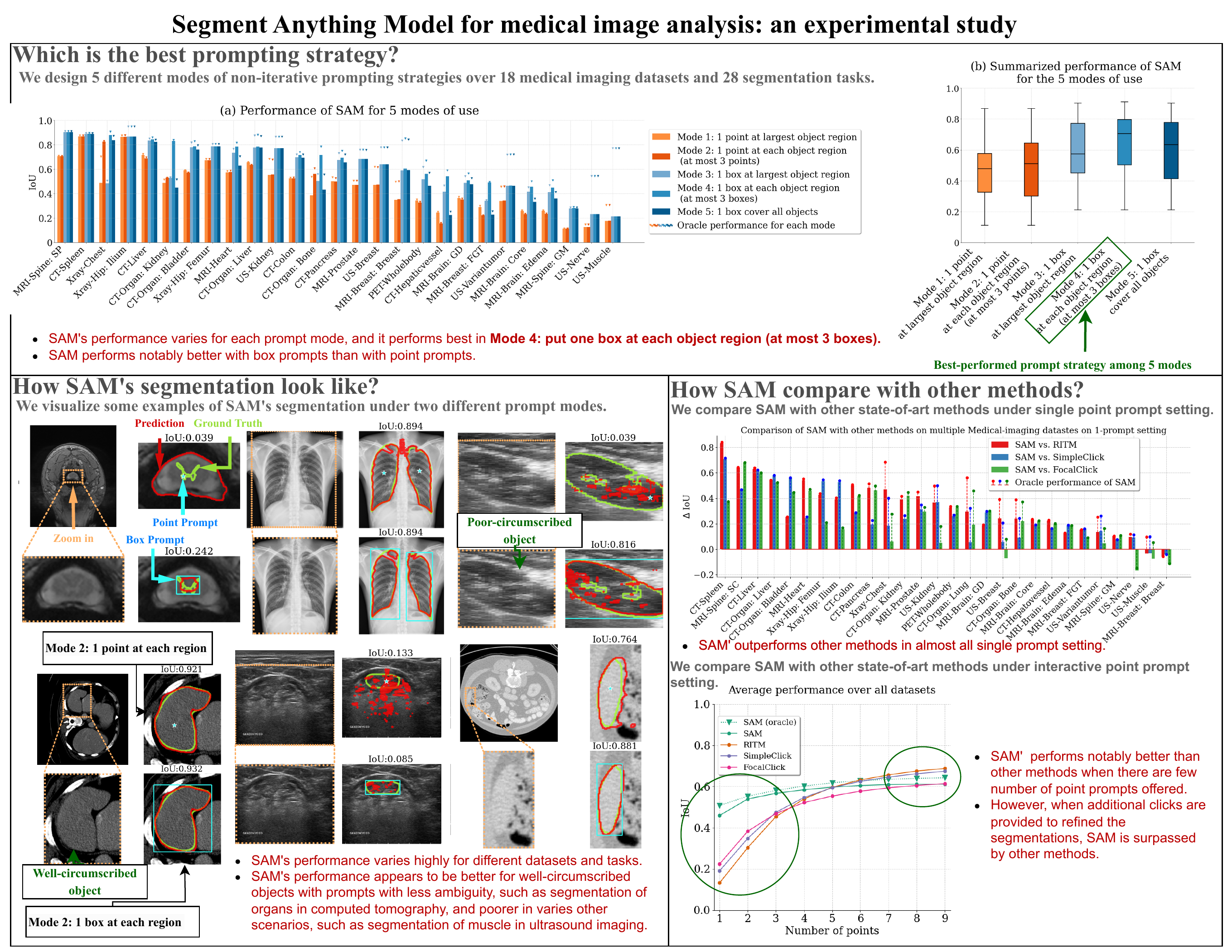}
    \caption{Graphical abstract of the work.}
    \label{fig:abs}
\end{figure*}

The graphical abstract of the work is shown in Fig. \ref{fig:abs}.

\subsection*{B. Examples of segmented results on all datasets}
The examples of segmented results over all 28 tasks are visualized in the following figures Fig. \ref{fig:exam1}-\ref{fig:exam5}. For each task, we display three examples with 25th, 50th, and 75th percentile of IoU.
For the PET/CT dataset, we are presenting PET information in the Red channel of the PNG file, and the CT information in the Green and Blue channels of the PNG file. 

\begin{figure*}
    \centering
    \includegraphics[page=1,scale=0.75]{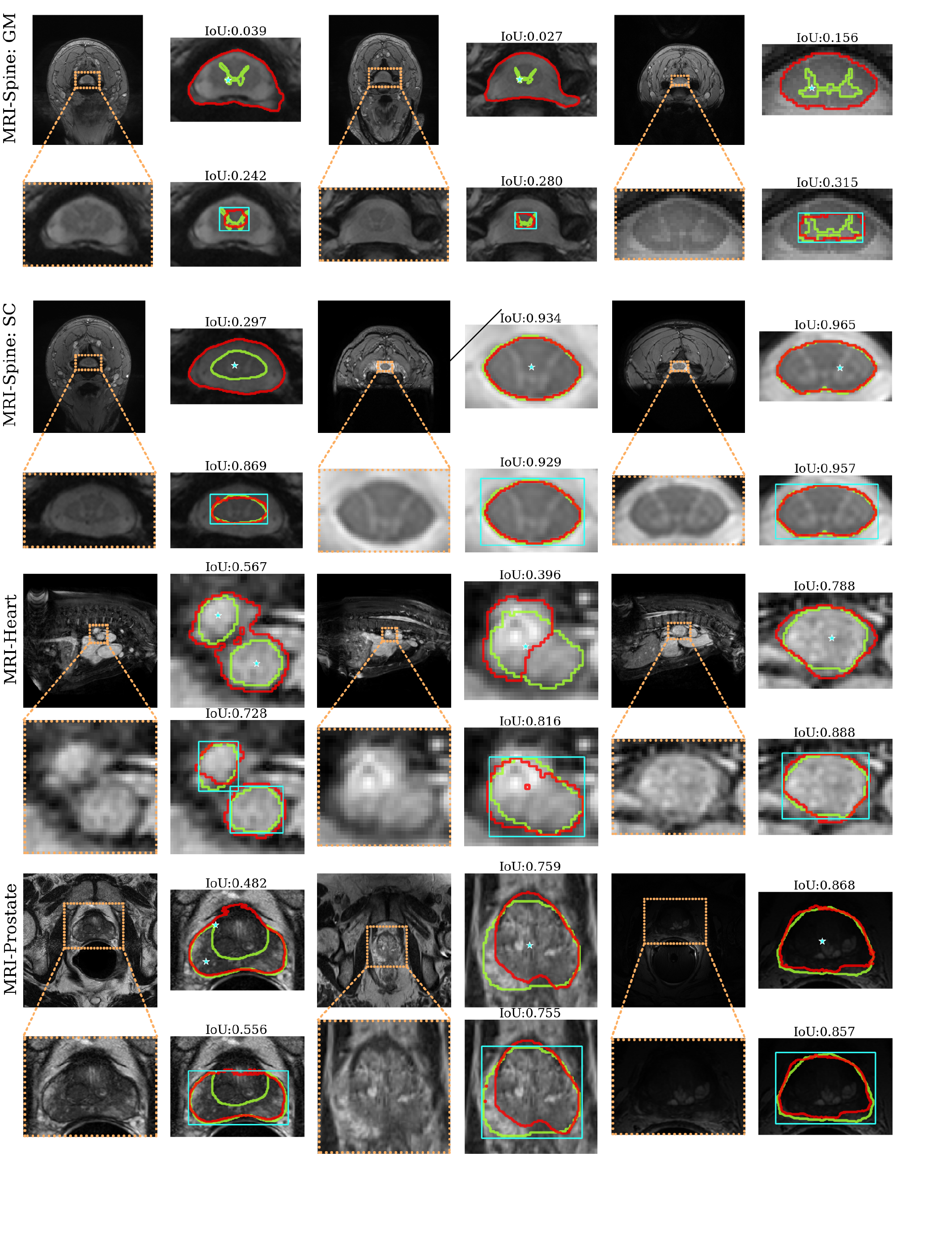}
    \caption{Visualization examples of SAM's prediction results (part1).}
    \label{fig:exam1}
\end{figure*}

\begin{figure*}
    \centering
    \includegraphics[page=2,scale=0.75]{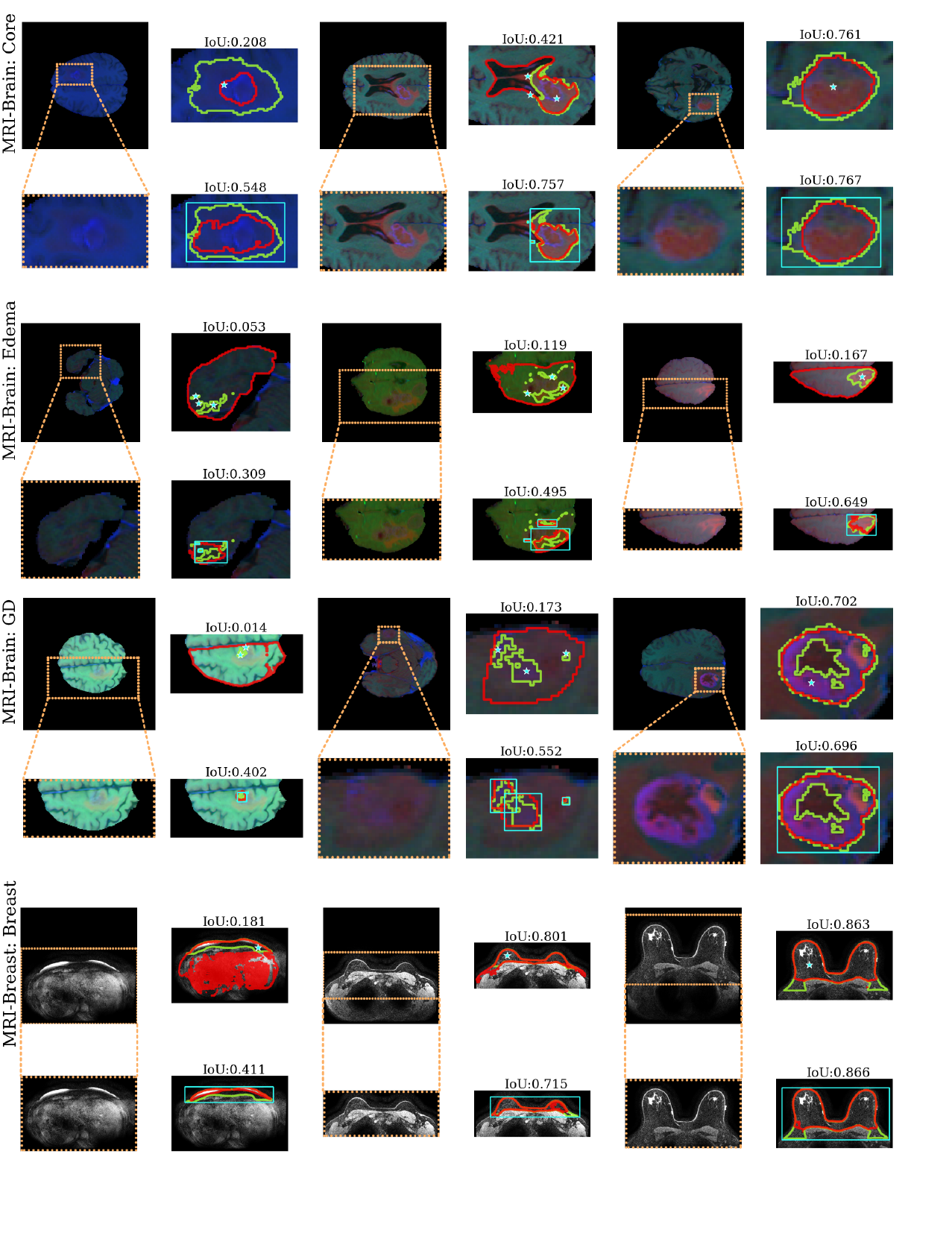}
    \caption{Visualization examples of SAM's prediction results (part2).}
    \label{fig:exam2}
\end{figure*}

\begin{figure*}
    \centering
    \includegraphics[page=1,scale=0.75]{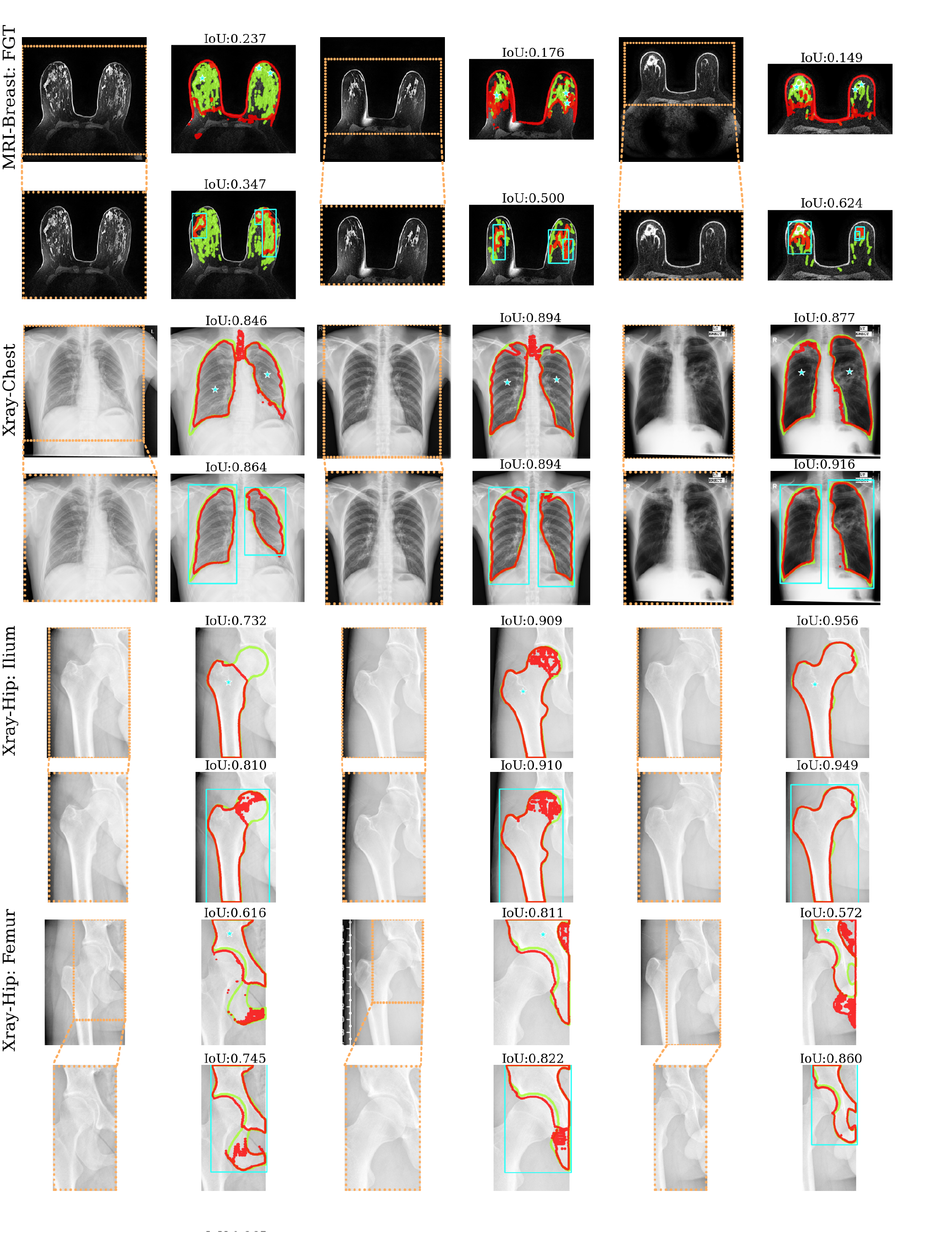}
    \caption{Visualization examples of SAM's prediction results (part3).}
    \label{fig:exam3}
\end{figure*}

\begin{figure*}
    \centering
    \includegraphics[page=2,scale=0.75]{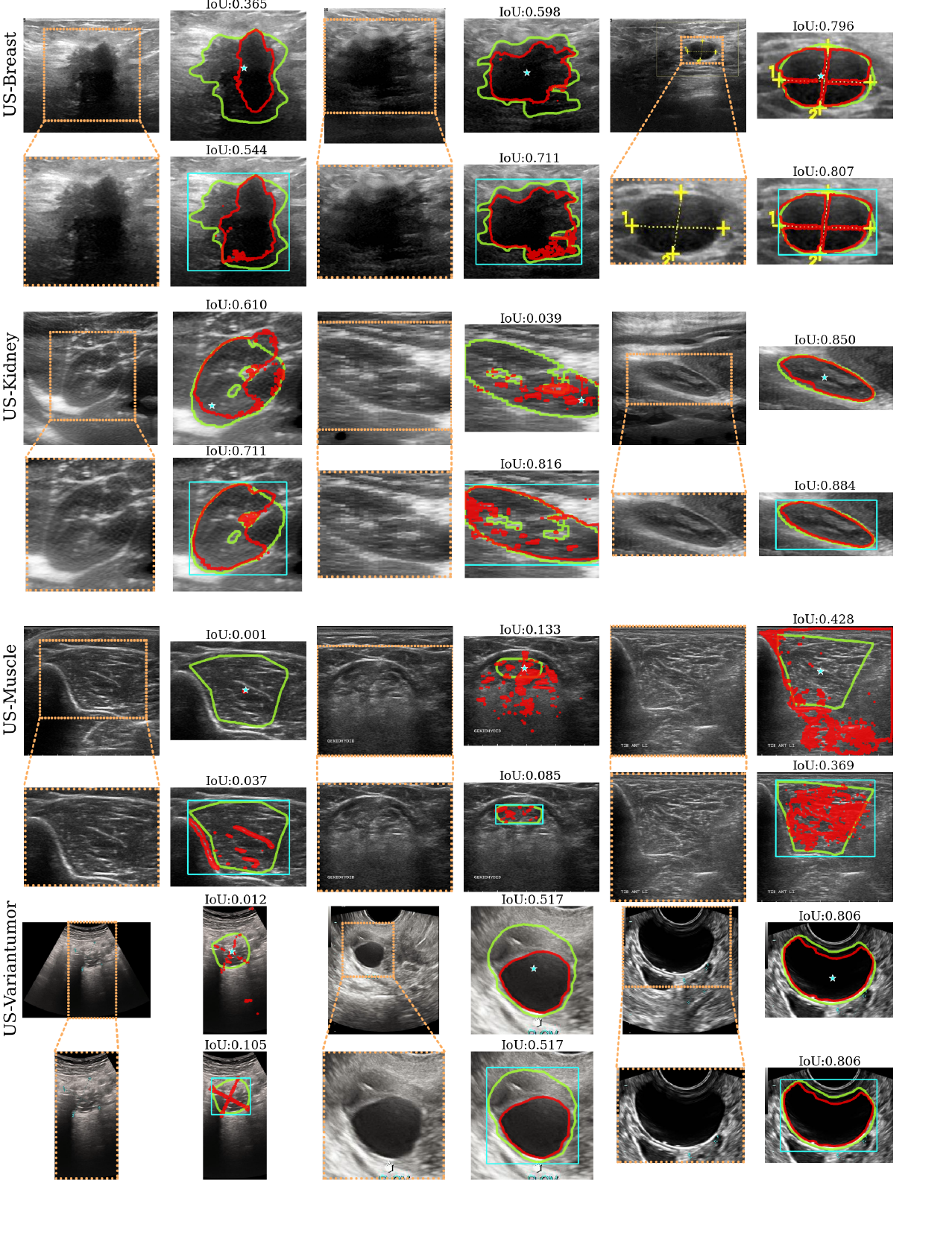}
    \caption{Visualization examples of SAM's prediction results (part4).}
    \label{fig:exam4}
\end{figure*}

\begin{figure*}
    \centering
    \includegraphics[page=1,scale=0.75]{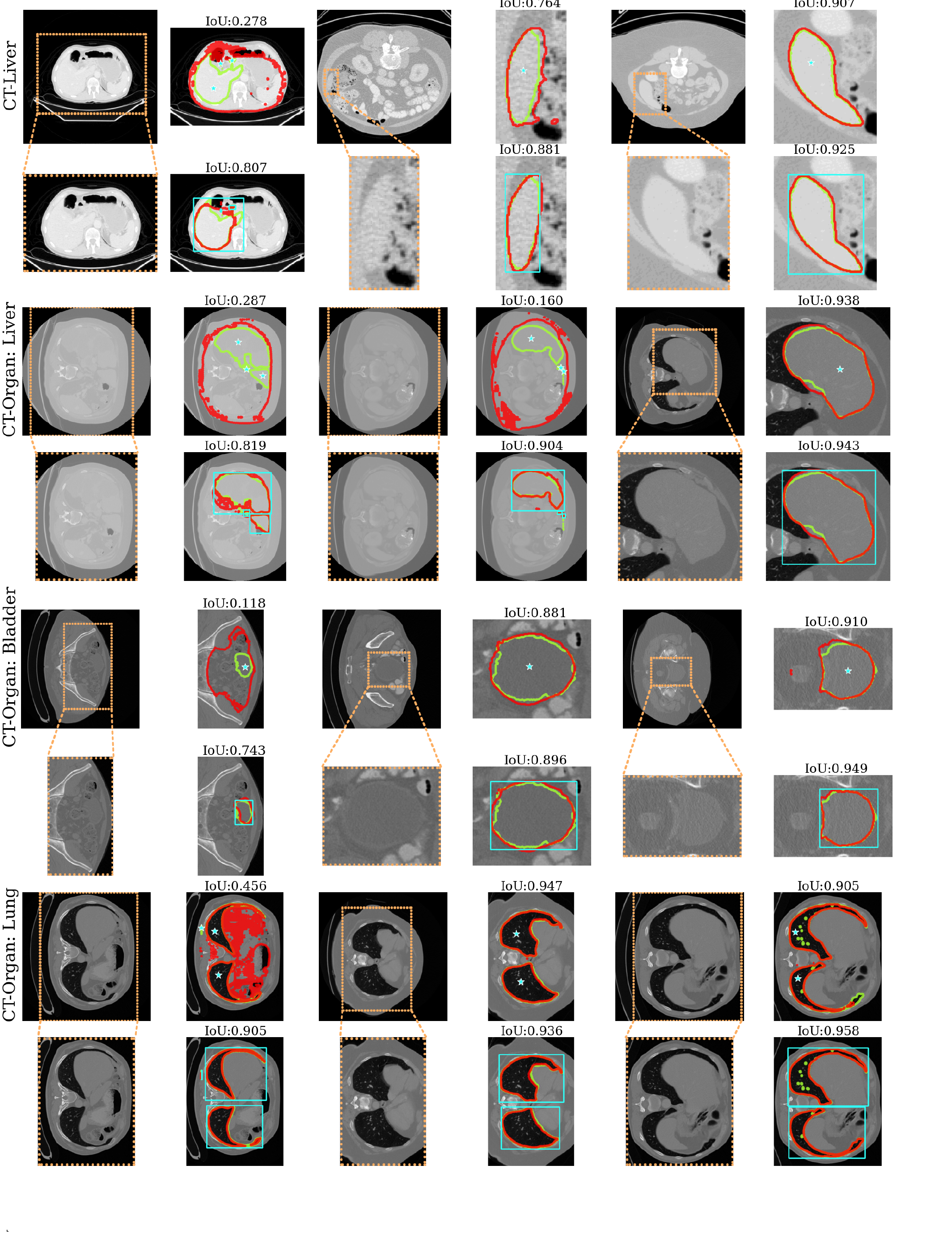}
    \caption{Visualization examples of SAM's prediction results (part5).}
    \label{fig:exam5}
\end{figure*}

\begin{figure*}
    \centering
    \includegraphics[page=2,scale=0.75]{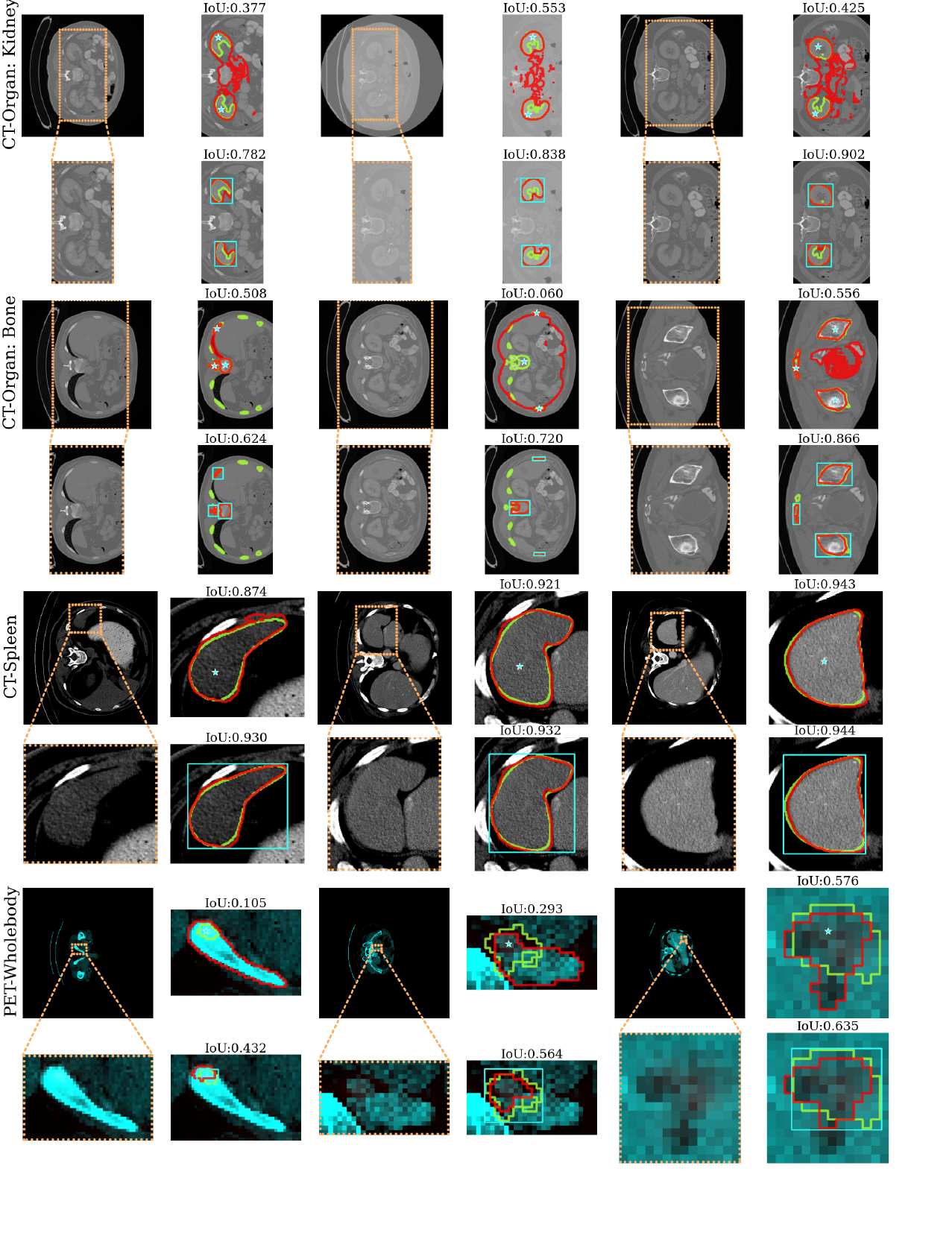}
    \caption{Visualization examples of SAM's prediction results.}
    \label{fig:exam5}
\end{figure*}

 \subsection*{ C. Performance of other competing methods}
\begin{figure*}
     \centering
     \includegraphics[width=1\linewidth]{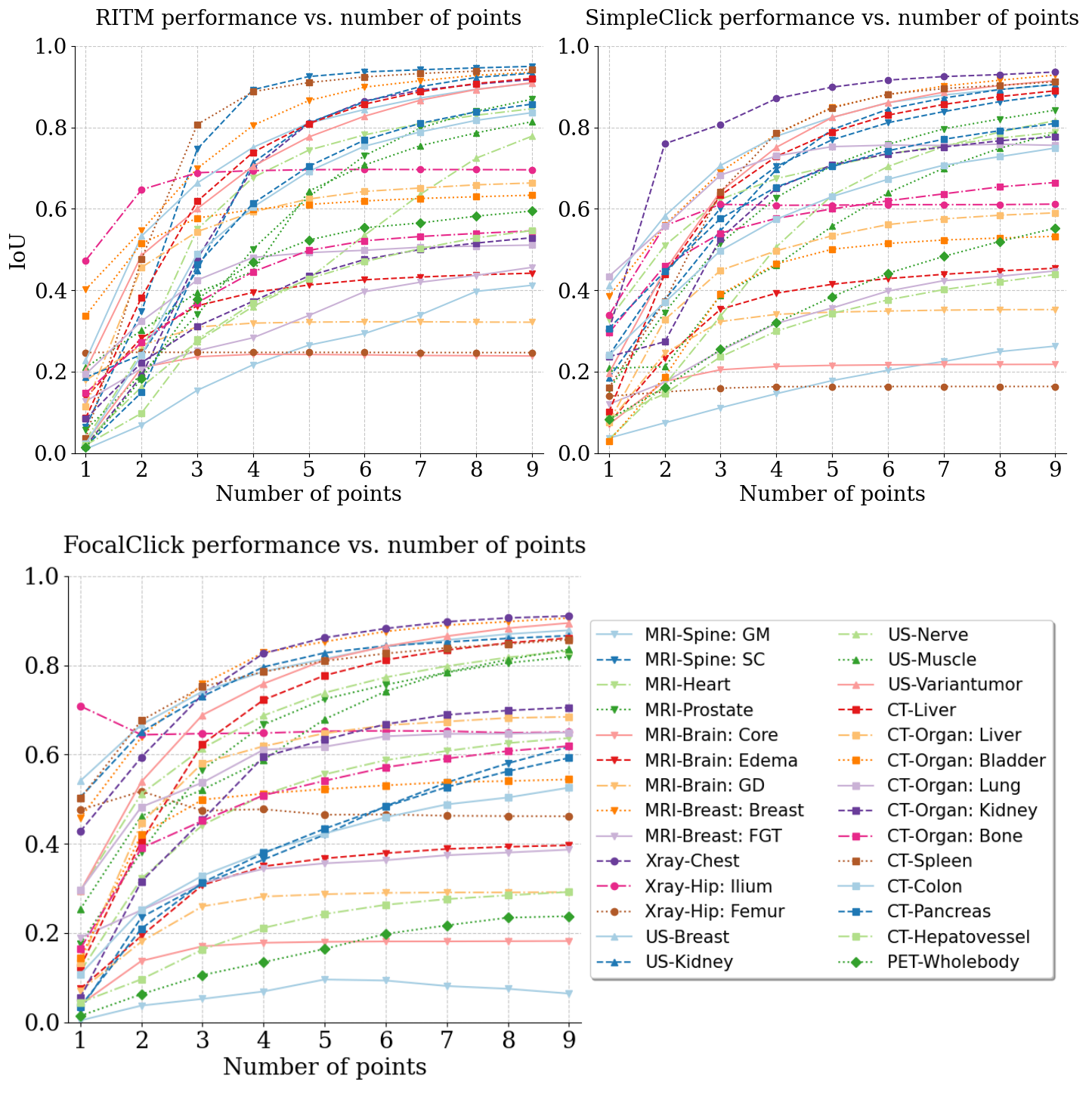}
     \caption{the detailed performance of RITM, SimpleClick, and FocalClick over each task under the interactive prompt setting.}
     \label{fig:other-methods}
 \end{figure*}

 The detailed performance of the other three competing methods over 28 tasks under the interactive prompt setting is shown in Figure \ref{fig:other-methods}.

\end{document}